\documentclass{article}
\usepackage{spconf,amsmath,bm,cite,graphicx,amssymb}
\usepackage{makecell,booktabs,arydshln}

\usepackage{color}
\usepackage{multirow}
\usepackage{etoolbox,siunitx}
\usepackage[noend]{algpseudocode}
\usepackage{algorithm}
\robustify\bfseries
\usepackage{balance}

\makeatletter
\def\adl@drawiv#1#2#3{%
        \hskip.5\tabcolsep
        \xleaders#3{#2.5\@tempdimb #1{1}#2.5\@tempdimb}%
                #2\z@ plus1fil minus1fil\relax
        \hskip.5\tabcolsep}
        
\newcommand{\cdashlinelr}[1]{%
  \noalign{\vskip\aboverulesep
           \global\let\@dashdrawstore\adl@draw
           \global\let\adl@draw\adl@drawiv}
  \cdashline{#1}
  \noalign{\global\let\adl@draw\@dashdrawstore
           \vskip\belowrulesep}}
           
\newcommand{\myhdashline}{%
  \noalign{\vskip\aboverulesep
           \global\let\@dashdrawstore\adl@draw
           \global\let\adl@draw\adl@drawiv}
  \hdashline
  \noalign{\global\let\adl@draw\@dashdrawstore
           \vskip\belowrulesep}}
\makeatother

\title{Sequence Transduction with Graph-based Supervision}
\name{Niko Moritz$^1$, Takaaki Hori$^1$, Shinji Watanabe$^2$, Jonathan Le Roux$^1$}
\address{$^1$Mitsubishi Electric Research Laboratories (MERL), Cambridge, MA, USA \\
$^2$Carnegie Mellon University, Pittsburgh, PA, USA} %

\begin{document}
\ninept
\setlength{\abovedisplayskip}{4pt}
\setlength{\belowdisplayskip}{4pt}
\setlength{\abovecaptionskip}{3pt}
\setlength{\belowcaptionskip}{-3pt}
\setlength{\textfloatsep}{5pt}

\maketitle

\begin{abstract}

The recurrent neural network transducer (RNN-T) objective plays a major role in building today's best automatic speech recognition (ASR) systems for production.
Similarly to the connectionist temporal classification (CTC) objective, the RNN-T loss uses specific rules that define how a set of alignments is generated to form a lattice for the full-sum training.
However, it is yet largely unknown if these rules are optimal and do lead to the best possible ASR results.
In this work, we present a new transducer objective function that generalizes the RNN-T loss to accept a graph representation of the labels, thus providing a flexible and efficient framework to manipulate training lattices, e.g., for studying different transition rules, implementing different transducer losses, or restricting alignments.
We demonstrate that transducer-based ASR with CTC-like lattice achieves better results compared to standard RNN-T, while also ensuring a strictly monotonic alignment, which will allow better optimization of the decoding procedure.
For example, the proposed CTC-like transducer achieves an improvement of 4.8\% on the test-other condition of LibriSpeech relative to an equivalent RNN-T based system.

\end{abstract}
\vspace{-0.1cm}
\section{Introduction}
\label{sec:intro}
\vspace{-0.1cm}

Two of the most popular neural network loss functions in automatic speech recognition (ASR) are the connectionist temporal classification (CTC) \cite{GravesFGS06} and Recurrent Neural Network Transducer (RNN-T) objectives \cite{Graves12}.
The CTC and RNN-T losses are designed for an alignment-free training of a neural network model to learn a mapping of a sequence of inputs (e.g., the acoustic features) to a typically shorter sequence of output labels (e.g., words or sub-word units).
While the CTC loss requires neural network outputs to be conditionally independent, the RNN-T loss provides an extension to train a neural network whose output frames are conditionally dependent on previous output labels.
In order to perform training without knowing the alignment between the input and output sequences, both loss types marginalize over a set of all possible alignments. Such alignments are derived from the supervision information (the sequence of labels) by applying specific instructions that define how the sequence of labels is expanded to adjust to the length of the input sequence. In both cases, such instructions include the usage of an additional blank label and transition rules that are specific to the loss type.
For example, in CTC, a blank label can be inserted at the beginning and end of the label sequence as well as between the ASR labels, and must be inserted between two similar ASR labels, and each label can get repeated as many times as necessary to match the input sequence length \cite{GravesFGS06}.
The RNN-T loss instead does not allow repetitions of an ASR label but the emission of multiple ASR labels per frame, which is possible because the neural network output has an additional dimensionality corresponding to a decoder state along which the training lattice is expanded \cite{Graves12}. The decoder state of an RNN-T model is obtained by the usage of an internal language model (LM), the predictor, where LM outputs are fused with the output of the encoder neural network using a joiner network \cite{Graves2013}.

Various prior studies in the literature have investigated using modifications of the CTC label transition rules such as Gram-CTC \cite{liu2017gramctc}, the automatic segmentation criterion (ASG) \cite{collobert2016wav2letter}, and the graph-based temporal classification (GTC) loss \cite{Moritz2021SemiSupervisedSR}.
GTC provides a generalization of CTC that accepts a graph representation of the labeling information, allowing for label transitions defined in a graph format.
Note that GTC has similarities to other recently proposed works on differentiable weighted finite state transducers such as GTN \cite{hannun2020dwfst} and k2 \cite{k2}, with the difference that GTN and k2 rely on automatic differentiation whereas gradients in GTC are manually computed.
However, while numerous works have focused on improving training, inference, and neural network architectures for RNN-T \cite{ImprovRNNT2019,Weinstein2020,SaonTA20,Liu2021ImprovingRT,BoyerESPNET21},
most studies that investigated altering the training lattice of transducer models focused on achieving a strictly monotonic alignment between the input and output sequences, and left other aspects of RNN-T, such as the emission of ASR labels over a single time frame, unaltered \cite{MonoRNNT19,JayAlignRest2021,zeyer2020new,Variani2020HybridAT}. Popular examples of RNN-T variants are the Recurrent Neural Aligner (RNA) \cite{RNA_Sak2017} and the Monotonic RNN-T (MonoRNN-T) \cite{MonoRNNT19} losses, whereby the main motivation for such variants is to better optimize the decoding process by using batching or vectorization techniques and to minimize delays \cite{seki19b_interspeech,MonoRNNT19}.

In this work, we propose the GTC-Transducer (GTC-T) objective, which extends GTC to conditional dependent neural network outputs similar to RNN-T.
GTC-T allows the user to define the label transitions in a graph format and by that to easily explore new lattice structures for transducer-based ASR.
Here, we propose to use a CTC-like lattice for training a GTC-T based ASR system, and compare results to a MonoRNN-T lattice type (also realized using GTC-T), standard RNN-T, as well as CTC. ASR results demonstrate that transducer-based ASR with a CTC-like graph outperforms all other loss types in terms of word error rates (WERs), especially when also using an external LM via shallow fusion.

\vspace{-0.1cm}
\section{GTC-Transducer}
\vspace{-0.1cm}

Let us consider a feature sequence $X$ of length $T'$ derived from a speech utterance, processed by a neural network to produce an output sequence of length $T$, potentially different from $T'$ due to downsampling. This output sequence contains a set of posterior probability distributions at every point, since the neural network is conditionally dependent on previous label outputs generated by the ASR system and therefore has different states producing multiple posterior probability distributions for the labels.
For example, $\bm \upsilon^{t,i}$ denotes the posterior probabilities for neural network state $i$ at time step $t$ and $\upsilon^{t,i}_k$ the posterior probability of output label $k$ for state $i$ at time $t$.
The Graph-based Temporal Classification-Transducer (GTC-T) objective function marginalizes over all possible label alignment sequences that are represented by the graph $\mathcal{G}$. Thus, the conditional probability for a given graph $\mathcal{G}$ is defined by the sum over all sequences of nodes in $\mathcal{G}$ of length $T$, which can be written as:
\begin{equation}
\label{eq:gtc}
    p(\mathcal{G}|X) = \sum_{\pi \in \mathcal{S}(\mathcal{G},T)} p(\pi | X),
\end{equation}
where $\mathcal{S}$ represents a search function that expands $\mathcal{G}$ to a lattice of length $T$ (not counting non-emitting start and end nodes), $\pi$ denotes a single node sequence and alignment path, and $p(\pi|X)$ is the posterior probability for the path $\pi$ given feature sequence $X$. 

We introduce a few more notations that will be useful to derive $p(\mathcal{G}|X)$. 
The nodes are sorted in a breadth-first search manner and indexed using $g=0,\dots,G+1$, where $0$ corresponds to the non-emitting start node and $G+1$ to the non-emitting end node. 
We denote by $l(g)$ the output symbol observed at node $g$, and by $W_{g,g'}$ and $I(g,g')$ the transition weight and the decoder state index on the edge connecting the nodes $g$ and $g'$.
Finally, we denote by $\pi_{t:t'}=(\pi_t,\dots,\pi_{t'})$ the node sub-sequence of $\pi$ from time index $t$ to $t'$.
Note that $\pi_0$ and $\pi_{T+1}$ correspond to the non-emitting start and end nodes $0$ and $G+1$.

In RNN-T, the conditional probabilities $p(\bm y | X)$ for a given label sequence $\bm y$ are computed efficiently by a dynamic programming algorithm, which is based on computing the forward and backward variables and combining them to compute $p(\bm y | X)$ at any given time $t$ \cite{Graves12}.
In a similar fashion, the GTC-T forward probability can be computed for $g=1,\dots,G$ using
\begin{equation}
\label{eq:grapctc_fw_probs_ext}
    \alpha_t (g) = \sum_{\substack{\pi \in \mathcal{S}(\mathcal{G},T):\\ \pi_{0:t} \in \mathcal{S}(\mathcal{G}_{0:g},t)}} \prod_{\tau=1}^t
    W_{\pi_{\tau-1}, \pi_{\tau}}
    \upsilon^{\tau ,I(\pi_{\tau-1},\pi_\tau)}_{l(\pi_{\tau})} ,
\end{equation}
where $\mathcal{G}_{0:g}$ denotes the sub-graph of $\mathcal{G}$ containing all paths from node $0$ to node $g$. 
The sum is taken over all possible $\pi$ whose sub-sequence up to time index $t$ can be generated in $t$ steps from the sub-graph $\mathcal{G}_{0:g}$.
Note that $\alpha_0(g)$ equals $1$ if $g$ corresponds to the start node and it equals $0$ otherwise.
The backward variable $\beta$ is computed similarly for $g=1,\dots,G$ 
using
\begin{equation}
\label{eq:grapctc_bw_probs_ext}
    \beta_t (g) = 
    \!\!\!\!\!\!\!\!\!\!\!\!\!\!\!\!\!
    \sum_{\substack{\pi \in \mathcal{S}(\mathcal{G},T):\\ \pi_{t:T+1} \in \mathcal{S}(\mathcal{G}_{g:G+1},T-t+1)}}
    \!\!\!\!\!\!\!\!\!\!\!\!\!\!\!\!\!
    W_{\pi_{T}, \pi_{T+1}}
    \prod_{\tau=t}^{T-1} 
    W_{\pi_{\tau}, \pi_{\tau+1}}
    \upsilon^{\tau+1 ,I(\pi_{\tau},\pi_{\tau+1})}_{l(\pi_{\tau+1})}\!\!\!\!,
\end{equation}
where $\mathcal{G}_{g:G+1}$ denotes the sub-graph of $\mathcal{G}$ containing all paths from node $g$ to node $G+1$. 
From the forward and backward variables at any $t$, the probability function $p(\mathcal{G}|X)$ can be computed using
\begin{equation}
\label{eq:fw_bw_ext_weights}
    p(\mathcal{G}|X) = \sum_{(g,g') \in \mathcal{G}} \alpha_{t-1}(g) W_{g, g'} \upsilon^{t ,I(g,g')}_{l(g')} \beta_t(g').
\end{equation}

For gradient descent training, the loss function
\begin{equation}
\label{eq:loss}
    \mathcal{L} = -\ln{p(\mathcal{G}|X)}
\end{equation}
must be differentiated with respect to the network outputs, which can be written as
\begin{equation}
\label{eq:deriv_ln_0_ext}
    -\frac{\partial \ln{ p(\mathcal{G}|X)} }{\partial \upsilon_{k}^{t,i}}  = -\frac{1}{p(\mathcal{G}|X)} \frac{\partial p(\mathcal{G}|X)}{\partial \upsilon_{k}^{t,i}},
\end{equation}
for any symbol $k \in \mathcal{U}$ and any decoder state $i \in \mathcal{I}$, where $\mathcal{U}$ denotes a set of all possible output symbols and $\mathcal{I}$ a set of all possible decoder state indices.
The derivative of $p(\mathcal{G}|X)$ with respect to $\upsilon_k^{t,i}$ can be written as
\begin{equation}
\label{eq:deriv_pGX_y_ext}
    \frac{\partial  p(\mathcal{G}|X)}{\partial \upsilon^{t,i}_{k}}  = \sum_{(g,g') \in \operatorname{\Phi}(\mathcal{G},k,i)} \alpha_{t-1}(g)  W_{g, g'} \beta_t(g'),
\end{equation}
where $\operatorname{\Phi}(\mathcal{G},k,i) = \{(g,g') \in \mathcal{G} : l(g') = k \wedge I(g,g') = i \}$ denotes the set of edges in $\mathcal{G}$ that correspond to decoder state $i$ and where label $k$ is observed at node $g'$.
To backpropagate the gradients through the softmax function of $\upsilon_k^{t,i}$, we need the derivative with respect to the unnormalized network outputs $h_k^{t,i}$ before the softmax is applied, which is
\begin{equation}
\label{eq:deriv_pGX_u_ext}
    -\frac{\partial \ln{{p}(\mathcal{G}|X)}}{\partial h^{t, i}_{k}} = - \sum_{k' \in \mathcal{U}} \frac{\partial \ln{{p}(\mathcal{G}|X)}}{\partial \upsilon^{t, i}_{k'} } \frac{\partial \upsilon^{t, i}_{k'}}{\partial h^{t , i}_{k}},.
\end{equation}
Finally, the gradients for the neural network outputs are
\begin{multline}
\label{eq:gradients}
    \hspace{-.3cm}-\frac{\partial \ln{ p(\mathcal{G}|X)} }{\partial h^{i,t}_k}  = \frac{ \upsilon^{t, i}_{k}}{{p}(\mathcal{G}|X)} \big(\sum_{(g,g') \in \operatorname{\Psi}(\mathcal{G},i)}\!\!\!\!\!\! \alpha_{t-1}(g) W_{g, g'} \upsilon^{t ,i}_{l(g')} \beta_t(g')\\ -\!\!\! \sum_{(g,g') \in \operatorname{\Phi}(\mathcal{G},k,i)}\!\!\! \alpha_{t-1}(g) W_{g, g'} \beta_t(g') \big) ,
\end{multline}
where $\operatorname{\Psi}(\mathcal{G},i) = \{(g,g') \in \mathcal{G} : I(g,g') = i \}$.
Eq.~(\ref{eq:gradients}) is derived by substituting (\ref{eq:deriv_pGX_y_ext}) and the derivative of the softmax function $ \partial \upsilon_{k'}^{t,i} / \partial h_k^{t,i} = \upsilon_{k'}^{t,i} \delta_{k k'}  -  \upsilon_{k'}^{t,i} \upsilon_k^{t,i} $ into (\ref{eq:deriv_pGX_u_ext})
and by using the fact that
\begin{align}
    -\sum_{k' \in \mathcal{U}} & \frac{\partial \ln{{p}(\mathcal{G}|X)}}{\partial \upsilon^{t, i}_{k'} } \upsilon^{t, i}_{k'} \delta_{k k'} 
    = -\frac{\partial \ln{{p}(\mathcal{G}|X)}}{\partial \upsilon^{t, i}_{k}} \upsilon^{t, i}_{k}  \nonumber \\
    &= -\frac{\upsilon^{t, i}_{k}}{{p}(\mathcal{G}|X)} \sum_{(g,g') \in \operatorname{\Phi}(\mathcal{G},k,i)} \alpha_{t-1}(g) W_{g, g'} \beta_t(g') ,
\end{align}
and that
\begin{align}
    \sum_{k' \in \mathcal{U}} & \frac{\partial \ln{{p}(\mathcal{G}|X)}}{\partial \upsilon^{t, i}_{k'}} \upsilon^{t, i}_{k'} \upsilon^{t, i}_{k} \nonumber \\
    &= \sum_{k' \in \mathcal{U}} \frac{\upsilon^{t, i}_{k'} \upsilon^{t, i}_{k}}{{p}(\mathcal{G}|X)} \sum_{(g,g') \in \operatorname{\Phi}(\mathcal{G},k',i)}
    \alpha_{t-1}(g) W_{g, g'} \beta_t(g') , \nonumber \\
    &=  \frac{ \upsilon^{t, i}_{k}}{{p}(\mathcal{G}|X)} \sum_{k' \in \mathcal{U}} \sum_{(g,g') \in \operatorname{\Phi}(\mathcal{G},k',i)} 
    \alpha_{t-1}(g) W_{g, g'} \upsilon^{t, i}_{k'} \beta_t(g') , \nonumber \\
    &=  \frac{ \upsilon^{t, i}_{k}}{{p}(\mathcal{G}|X)} \sum_{(g,g') \in \operatorname{\Psi}(\mathcal{G},i)} \alpha_{t-1}(g) W_{g, g'} \upsilon^{t ,i}_{l(g')} \beta_t(g') .
\end{align}

The GTC-T loss is implemented in CUDA as an extension for pytorch to make it efficient.

\vspace{-0.1cm}
\section{Graph Topology}
\vspace{-0.1cm}

\begin{figure}[t]
  \centering
  \centerline{\includegraphics[width=0.85\linewidth]{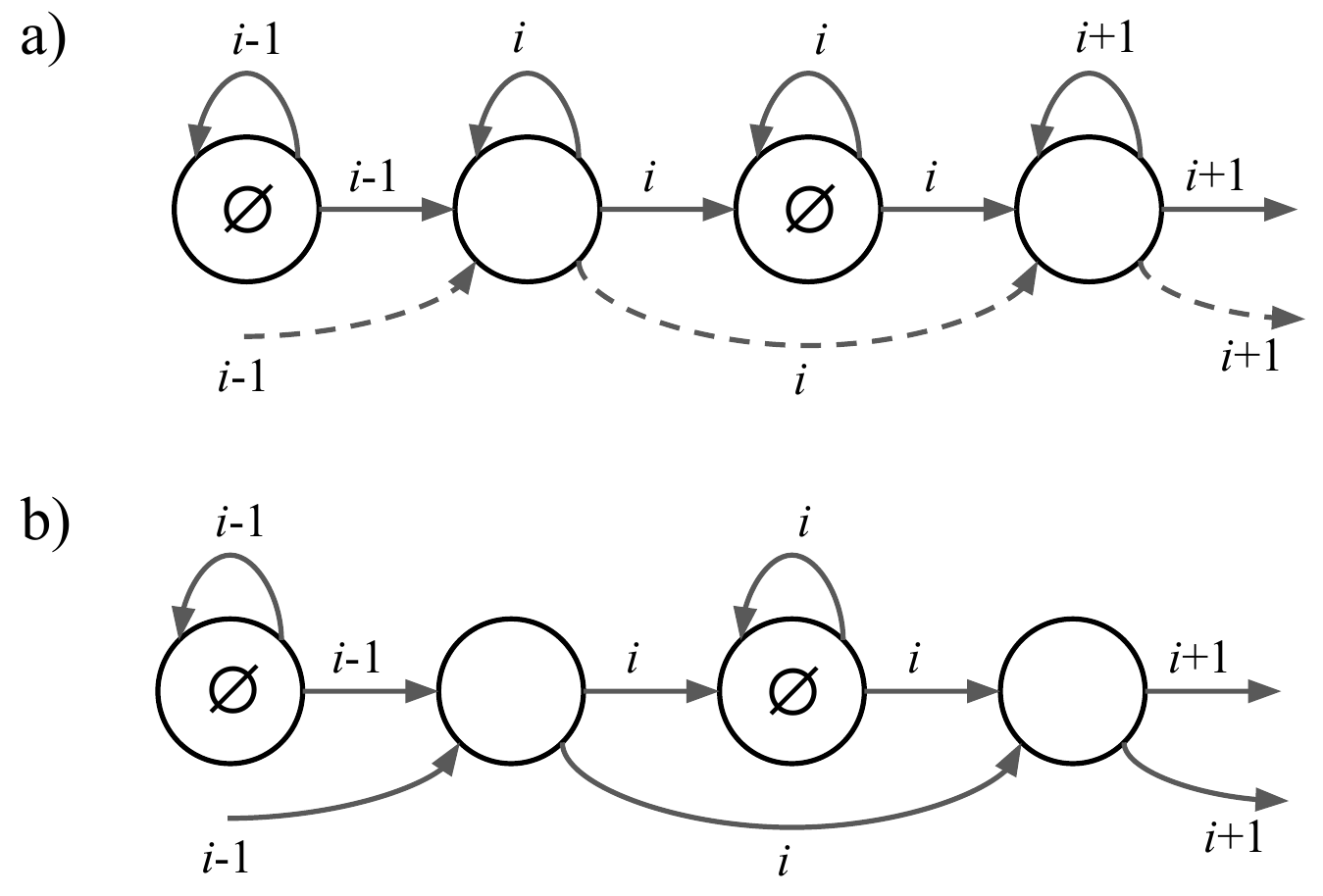}} \vspace{-2mm}
  \caption{Graph topologies for GTC-T training: a) CTC-like graph, b) MonoRNN-T graph. The neural network state are indicated by $i$, $\varnothing$ denotes the blank label, and empty nodes (circles) represent an ASR label. } %
\label{fig:graph_topos}
\end{figure}

The GTC-T objective allows the usage of different graph topologies for constructing the training lattice. In this work, we test two different graph types as shown in Fig.~\ref{fig:graph_topos}, where arrows correspond to edges and circles to nodes at which either a blank label, denoted by $\varnothing$, or an ASR label (empty circles) is emitted.
Neural network states are indicated using $i$ and reside on the edges of the graph.

Graph type a) of Fig.~\ref{fig:graph_topos} corresponds to a CTC-like topology, i.e., the graph can insert blanks between ASR labels following the CTC transition rules \cite{GravesFGS06} and each label can get repeated as many times as necessary to match the target sequence length. Dashed lines correspond to optional edges, which are used to account for the fact that blank nodes can be skipped unless the ASR labels of two consecutive nodes are the same, see CTC rules \cite{GravesFGS06}.

Graph type b) of Fig.~\ref{fig:graph_topos} corresponds to a MonoRNN-T loss type \cite{MonoRNNT19}. The main difference of MonoRNN-T to standard RNN-T is that multiple ASR outputs per time frame are not permitted, i.e., a strictly monotonic alignment between the input sequence and the output sequence is enforced.

Note that in order to ensure that the probabilities of all alignments in training will sum to one at most, the graph topology for GTC-T training must be carefully selected. In general, this means that the graph should be deterministic and that the posterior probabilities of all outgoing edges of a node should be derived from the same distribution, i.e., generated by using the same neural network state. Also note that all weights $W_{g,g'}$ of the graphs are set to one in this work.

\vspace{-0.1cm}
\section{Decoding Algorithm}
\label{sec:decoding}
\vspace{-0.1cm}

The beam search algorithm for GTC-T with a CTC-like graph is shown as pseudo-code in Algorithm~\ref{alg:decode}. The algorithm is inspired by a frame-synchronous CTC prefix beam search algorithm \cite{HannunMJN14}. In this notation, $\ell$ corresponds to a prefix sequence, the prefix probability is separated into $p^t_\text{nb}(\ell)$ and $p^t_\text{b}(\ell)$ for ending with in blank (b) or not ending in blank (nb) at time step $t$, and $\theta_1$ and $\theta_2$ are used as thresholds for pruning the set of posterior probabilities locally and for score-based pruning of the set of prefixes/hypotheses.
More specifically, function $\textsc{Prune}(\Omega_\text{next},p_\text{asr},P,\theta_2)$ performs two pruning steps. First, the set of hypotheses residing in $\Omega_\text{next}$ is limited to the $P$ best hypotheses using the ASR scores $p_\text{asr}$, then any ASR hypothesis whose ASR score is less than $\log p_\text{best} - \theta_2$ is also removed from the set, where $p_\text{best}$ denotes the best prefix ASR score in the set.
The posterior probabilities $\bm \upsilon^{t,i}$ are generated by the neural network using $\textsc{NNet}(X,\ell,t)$, where $X$ represents the input feature sequence, and $i$ denotes the neural network state that depends on prefix $\ell$.
The posterior probability of ASR label $k$ is denoted by $\upsilon^{t,i}_k$ and the neural network output $\upsilon^{t,j}_k$ for state $j$, cf.\ line 24 to 26, is associated with the prefix $\ell_+$.
Furthermore, $\alpha$ and $\beta$ are the LM and label insertion bonus weights \cite{HannunMJN14,MoritzHLR20,MoritzHR19c} and $|\ell|$ denotes the sequence length of prefix $\ell$. The $\varnothing$ symbol represents the blank label and $\langle \text{sos} \rangle$ a start of sentence symbol. 

\begin{algorithm}[t]
\caption{Beam search for GTC-T with a CTC-like graph. } \label{alg:decode}
\begin{algorithmic}[1]
\State $\ell \gets (\langle\text{sos}\rangle,)$
\State $p_\text{nb}^0(\ell) \gets 0$, $p_\text{b}^0(\ell) \gets 1$
\State $\Omega_\text{pruned} \gets \{\ell\}$, $\Omega_\text{prev} \gets \{\}$ 
\For{$t=1,\dots,T$}
  \State $\Omega_\text{next} \gets \{\}$
  \For{$\ell$ \textbf{in} $\Omega_\text{pruned}$}
    \State $\bm \upsilon^{t,i} \gets \textsc{NNet}(X,\ell,t)$
    \State $C \gets \{k \textbf{ for } k \textbf{ in } \mathcal{U} \textbf{ if } \upsilon_k^{t,i} > \theta_1$\}
    \State add $\varnothing$ to $C$
    \For{$k$ \textbf{in} $C$}
      \If{$k = \varnothing$}
        \State $p_\text{b}^t(\ell) \gets \upsilon_k^{t,i} (p_\text{b}^{t-1}(\ell) + p_\text{nb}^{t-1}(\ell))$
        \If{$|\ell|>1$ \textbf{and} $\ell_\text{end}$ \textbf{not in} $C$}
          \State $p_\text{nb}^t(\ell) \gets \upsilon_{\ell_\text{end}}^{t,i} p_\text{nb}^{t-1}(\ell)$
        \EndIf
        \State add $\ell$ to $\Omega_\text{next}$
      \Else
        \State $\ell_+ \gets \ell + (k,)$
        \If{$k = \ell_\text{end}$}
          \State $p_\text{nb}^t(\ell_+) \gets \upsilon_k^{t,i} p_\text{b}^{t-1}(\ell)$
          \State $p_\text{nb}^t(\ell) \gets \upsilon_k^{t,i} p_\text{nb}^{t-1}(\ell)$
        \Else
          \State $p_\text{nb}^t(\ell_+) \gets \upsilon_k^{t,i} (p_\text{b}^{t-1}(\ell) + p_\text{nb}^{t-1}(\ell))$
        \EndIf
        \If{$\ell_+$ \textbf{not in} $\Omega_\text{pruned}$ \textbf{and} $\ell_+$ \textbf{in} $\Omega_\text{prev}$}
          \State $\bm \upsilon^{t,j} \gets \textsc{NNet}(X,\ell_+,t)$
          \State $ p_\text{b}^t(\ell_+) \gets \upsilon_{\varnothing}^{t,j}(p_\text{b}^{t-1}(\ell_+)+p_\text{nb}^{t-1}(\ell_+))$
          \State $p_\text{nb}^t(\ell_+) \gets \upsilon_{k}^{t,j} p_\text{nb}^{t-1}(\ell_+)$ 
        \EndIf
        \State add $\ell_+$ to $\Omega_\text{next}$
      \EndIf
    \EndFor
  \EndFor
  \For{$\ell$ \textbf{in} $\Omega_\text{next}$}
    \State $p_\text{asr}(\ell) \gets (p_\text{b}^{t}(\ell) + p_\text{nb}^{t}(\ell)) p_\text{LM}(\ell)^\alpha |\ell|^\beta$
  \EndFor
  \State $\Omega_\text{pruned} \gets \textsc{Prune}(\Omega_\text{next},p_\text{asr},P,\theta_2)$
  \State $\Omega_\text{prev} \gets \Omega_\text{next}$
\EndFor
\end{algorithmic}
\end{algorithm}

\vspace{-0.1cm}
\section{Experiments}
\vspace{-0.1cm}

\subsection{Setup}

We use the HKUST \cite{hkust} and the LibriSpeech \cite{librispeech} ASR benchmarks for evaluation.
HKUST is a corpus of Mandarin telephone speech recordings with more than 180 hours of transcribed speech data, and LibriSpeech comprises nearly 1k hours of read English audio books.

The ASR systems of this work are configured to first extract 80-dimensional log-mel spectral energies plus 3 extra features for pitch information \cite{watanabe2018espnet}. The derived feature sequence is processed by a 2-layer VGG neural network \cite{HoriWZC17}, which downsamples the sequence of features to a frame rate of 60~ms, before being fed into a Conformer encoder architecture \cite{gulati2020conformer}.
The encoder neural network is composed of 12 Conformer blocks, where each block includes a self-attention layer, a convolution module, and two Macaron-like feed-forward neural network modules \cite{gulati2020conformer}.
In addition, the input to each component of the Conformer block is layer normalized and dropout is applied to the output of several neural network layers similar to \cite{Moritz_DCN}.
Hyperparameters of the Conformer encoder are similar to \cite{Moritz_DCN}, i.e., $d_\mathrm{model}=256$, $d_\mathrm{ff}=2048$, $d_h=4$, and $E=12$ for HKUST, while $d_\mathrm{model}$ and $d_h$ are increased to $512$ and $8$ for LibriSpeech.
For the CTC model, the output of the encoder neural network is projected to the number of output labels (including the blank label) using a linear layer and a softmax function to derive a probability distribution over the labels.
For the GTC-T and RNN-T loss types, two additional neural network components are used, the prediction network and the joiner network. The prediction network, which consists of a single long short-term memory (LSTM) neural network and a dropout layer, acts like a language model and receives as an input the previously emitted ASR labels (ignoring the blank label), which are converted into an embedding space.
The joiner network combines the sequence of encoder frames and the prediction neural network outputs using a few linear layers and a tanh activation function. Finally, a softmax is used to produce a probability distribution for the labels.

The Adam optimizer with $\beta_1=0.9$, $\beta_2=0.98$, $\epsilon=10^{-9}$, and learning rate scheduling similar to \cite{VaswaniSPUJGKP17} with 25000 warmup steps is applied for training.
The learning rate factor and the maximum number of training epochs are set to $1.0$ and $50$ for HKUST and to $5.0$ and $100$ for LibriSpeech. 
SpecAugment is used for all experiments \cite{ParkCZC19}.
A task-specific LSTM-based language model (LM) is trained using the official training text data of each ASR task \cite{librispeech,hkust} and employed via shallow fusion during decoding whenever indicated. For HKUST, the LM consists of 2 LSTM layers with 650 units each. For LibriSpeech, 4 LSTM layers with 2048 units each are used instead. For LibriSpeech, we also test the effect of a strong Transformer-based LM (Tr-LM) with 16 layers. ASR output labels consist of a blank token plus 5,000 subword units obtained by the SentencePiece method \cite{KudoR18} for LibriSpeech or of a blank token plus 3,653 character-based symbols for the Mandarin HKUST task.

Note that, in the following results section, greedy search is taking the underlying loss types into account, i.e., label sequences are collapsed according to the lattice topologies. The beam search method for RNN-T is based on the standard decoding algorithm proposed by Graves \cite{Graves12}, and the GTC-T beam search with a CTC-like graph is explained in Section \ref{sec:decoding}.

\vspace{-0.1cm}
\subsection{Results}

\begin{table}[tb]
  \caption{ HKUST ASR results. CTC15 denotes parameter initialization using the snapshot after 15 epochs of CTC training, BS10 denotes beam search with beam size 10, and joint indicates multi-objective training using RNN-T and CTC. }
  \label{tab:results_hkust}
  \centering
     \sisetup{table-format=2.1,round-mode=places,round-precision=1,table-number-alignment = center,detect-weight=true,detect-inline-weight=math}
  \resizebox{.97\linewidth}{!}
  {\setlength{\tabcolsep}{4pt}
  \begin{tabular}{cccccSS}
  \toprule
  \multicolumn{3}{c}{Training} & \multicolumn{2}{c}{Decoding} & \multicolumn{2}{c}{WER [\%]} \\
  Loss & Graph & Init & Search & LM & {train-dev} & {dev} \\
  \cmidrule(lr){1-3}\cmidrule(lr){4-5}\cmidrule(lr){6-7}
  CTC & - & - & greedy & - & 21.3 & 21.6 \\
  CTC & - & - & BS10 & LSTM & 20.3 & 20.9 \\
  
  RNN-T & - & - & BS10 & - & 22.7 & 23.1 \\
  RNN-T & - & joint & BS10 & - & 21.5 & 22.2 \\
  RNN-T & - & joint & BS10 & LSTM & 21.3 & 22.2 \\

  GTC-T & CTC-like & - & greedy & - & 24.2 & 24.4 \\
  GTC-T & CTC-like & CTC15 & greedy & - & 21.4 & 22.1 \\
  GTC-T & MonoRNN-T & CTC15 & greedy & - & 21.9 & 22.7 \\
  GTC-T & CTC-like & CTC15 & BS10 & - & 20.9 & 21.8 \\
  GTC-T & CTC-like & CTC15 & BS10 & LSTM & 20.8 & 21.7 \\
  
\bottomrule
  \end{tabular}}
\end{table}

ASR results for the CTC, RNN-T, and GTC-T losses on the HKUST benchmark are shown in Table~\ref{tab:results_hkust}.
Joint CTC / RNN-T training \cite{BoyerESPNET21} as well as parameter initialization for GTC-T training via CTC pre-training greatly improves ASR results for both RNN-T and GTC-T based models.
Note that CTC-based initialization only affects parameters of the encoder neural network, while parameters of the prediction and joiner network remain randomly initialized. We leave better initialization of such model components to future work.
The ASR results demonstrate that for GTC-T training the usage of a CTC-like graph performs better compared to a MonoRNN-T graph. In addition, the GTC-T model outperforms the results of the RNN-T model by 0.5\% on the HKUST dev test set.
While the usage of an LM via shallow fusion did not help to improve word error rates (WERs) for HKUST much in general, the CTC-based ASR results benefit the most with improvements between 0.7\% and 1.0\%.
For HKUST, the CTC system also outperformed both the RNN-T as well as the GTC-T systems. We suspect the reasons for it is that RNN-T models are known to be data hungry \cite{Li2020OnTC} and the training data size is probably too small to show the full potential of transducer-based ASR systems.

ASR results on the larger LibriSpeech dataset are shown in Table~\ref{tab:results_libri}, where RNN-T as well as GTC-T clearly outperform CTC results. For example, GTC-T with a CTC-like graph, CTC-based initialization, a Transformer-based LM, and a beam size of 30 for decoding achieves a WERs of 5.9\% for the test-other conditions of LibriSpeech. This is 0.9\% better compared to the best CTC results despite using a strong LM and a generous beam size. The GTC-T results are also 0.3\% better compared to the best RNN-T results of this work.
In addition, similar to the HKUST experiments, it can be noticed that GTC-T with a CTC-like graph obtains better results than using the MonoRNN-T graph.
However, the results of Table~\ref{tab:results_libri} also demonstrate that parameter initialization of the encoder neural network is particularly important for GTC-T training, and without initialization the training converges more slowly.
We also conducted extra experiments by training the GTC-T model for more epochs when model parameters are not initialized (results are not shown in the table), which further improved the ASR results.
In comparison, RNN-T model training converged faster in our experiments and extra training did not help improve results any further, which supports the assumption that initialization is particularly important for the GTC-T based training.
We can also notice from the results that, for LibriSpeech, the RNN-T model performs better than GTC-T when no external LM is used. Conversely, this also indicates that the GTC-T system can make better use of an extra LM via shallow fusion, but the investigation of this finding, especially with respect to personalization issues of transducer-based ASR systems, remains for future work. 

\begin{table}[tb]
  \caption{ WERs [\%] for LibriSpeech. CTC20 indicates parameter initialization (Init) from epoch 20 of CTC training and CTC under Init denotes parameter initialization from a fully trained CTC model. }
  \label{tab:results_libri}
  \centering
     \sisetup{table-format=2.1,round-mode=places,round-precision=1,table-number-alignment = center,detect-weight=true,detect-inline-weight=math}
  \resizebox{.97\linewidth}{!}
  {\setlength{\tabcolsep}{4pt}
  \begin{tabular}{cccccSSSS}
  \toprule
  \multicolumn{3}{c}{Training} & \multicolumn{2}{c}{Decoding} & \multicolumn{2}{c}{dev} & \multicolumn{2}{c}{test} \\
  Loss & Graph & Init & Search & LM & {clean} & {other} & {clean} & {other} \\
  \cmidrule(lr){1-3}\cmidrule(lr){4-5}\cmidrule(lr){6-7}\cmidrule(lr){8-9}
  CTC & - & - & greedy & - & 4.9 & 12.0 & 5.0 & 11.7 \\
  CTC & - & - & BS10 & LSTM & 2.8 & 7.2 & 2.9 & 7.3 \\
  CTC & - & - & BS30 & LSTM & 2.7 & 7.1 & 2.9 & 7.1 \\
  CTC & - & - & BS10 & Tr. & 2.6 & 6.9 & 2.8 & 6.9 \\
  CTC & - & - & BS30 & Tr. & 2.5 & 6.8 & 2.5 & 6.8 \\
  
  RNN-T & - & - & greedy & - & 3.2 & 8.0 & 3.3 & 8.2 \\
  RNN-T & - & - & BS10 & - & 3.0 & 7.8 & 3.1 & 8.0 \\
  RNN-T & - & joint & BS10 & - & 2.9 & 7.8 & 3.1 & 7.8 \\
  RNN-T & - & joint & BS10 & LSTM & 2.4 & 6.6 & 2.7 & 6.7 \\
  RNN-T & - & joint & BS30 & LSTM & 2.4 & 6.3 & 2.6 & 6.4 \\
  RNN-T & - & joint & BS10 & Tr. & 2.4 & 6.8 & 2.7 & 6.5 \\
  RNN-T & - & joint & BS30 & Tr. & 2.3 & 6.2 & 2.5 & 6.2 \\

  GTC-T & MonoRNN-T & - & greedy & - & 5.3 & 13.2 & 5.4 & 13.5 \\ %
  GTC-T & MonoRNN-T & CTC20 & greedy & - & 4.1 & 10.3 & 4.2 & 10.5 \\ %
  GTC-T & CTC-like & - & greedy & - & 4.3 & 11.0 & 4.5 & 11.2 \\
  GTC-T & CTC-like & - & BS10 & - & 4.2 & 10.4 & 4.3 & 10.6 \\
  GTC-T & CTC-like & CTC20 & BS10 & - & 3.4 & 8.8 & 3.6 & 9.0 \\
  GTC-T & CTC-like & CTC & BS10 & - & 3.2 & 8.4 & 3.4 & 8.5 \\
  GTC-T & CTC-like & CTC & BS10 & LSTM & 2.4 & 6.1 & 2.7 & 6.2 \\
  GTC-T & CTC-like & CTC & BS30 & LSTM & 2.4 & 6.0 & 2.6 & 6.2 \\
  GTC-T & CTC-like & CTC & BS10 & Tr. & 2.3 & 6.0 & 2.5 & 6.0 \\
  GTC-T & CTC-like & CTC & BS30 & Tr. & 2.3 & 5.8 & 2.5 & 5.9 \\
  
\bottomrule
  \end{tabular}}
\end{table}

\vspace{-0.1cm}
\section{Conclusions}
\vspace{-0.1cm}

The proposed GTC-T loss provides a general framework for training transducer-based ASR models, where instructions for generating the training lattice are defined in graph format.
We found that GTC-T with a CTC-like lattice outperforms standard RNN-T in terms of WERs, while also omitting a practical issue of RNN-T by not permitting repeated ASR outputs per time frame, which allows for better optimization of the decoding procedure. On LibriSpeech, the proposed CTC-like transducer ASR system achieved WERs of 2.5\% (test-clean) and 5.9\% (test-other), which is a relative improvement of almost 5\% compared to standard RNN-T for the test-other condition. 

\balance
\bibliographystyle{IEEEtran}
\bibliography{refs}

\end{document}